\documentclass[runningheads]{llncs}

\usepackage{iciap}

\usepackage{iciapabbrv}
\usepackage{multirow}

\usepackage{graphicx}
\usepackage{booktabs}

\makeatletter
\newif\if@restonecol
\makeatother

\usepackage[linesnumbered,ruled,vlined]{algorithm2e}

\usepackage{algpseudocode}
\usepackage{amsmath}

\usepackage[accsupp]{axessibility}  

\usepackage[pagebackref,breaklinks,colorlinks,citecolor=iciapblue]{hyperref}

\usepackage{orcidlink}
\newcommand{\COMMENT}[1]{\textcolor{blue}{}}

\begin{document}

\title{A Semantically-Aware Relevance Measure for Content-Based Medical Image Retrieval Evaluation} 

\titlerunning{Abbreviated paper title}

\author{Xiaoyang Wei\orcidlink{0009-0003-8861-0427} \and
Camille Kurtz\orcidlink{0000-0001-9254-7537} \and
Florence Cloppet\orcidlink{0000-0002-4807-299X}}

\authorrunning{F.~Author et al.}

\institute{Universit\'e Paris Cit\'e, LIPADE, F-75006, Paris, France\\
\email{firstname.lastname@u-paris.fr}}

\maketitle

\begin{abstract}
Performance evaluation for Content-Based Image Retrieval (CBIR) remains a crucial but unsolved problem today especially in the medical domain. 
Various evaluation metrics have been discussed in the literature to solve this problem. 
Most of the existing metrics (e.g., precision, recall) are adapted from classification tasks which require manual labels as ground truth. 
However, such labels are often expensive and unavailable in specific thematic domains. 
Furthermore, medical images are usually associated with (radiological) case reports or annotated with descriptive captions in literature figures, such text contains information that can help to assess CBIR.
Several researchers have argued that the medical concepts hidden in the text can serve as the basis for CBIR evaluation purpose. 
However, these works often consider these medical concepts as independent and isolated labels while in fact the subtle relationships between various concepts are neglected. 
In this work, we introduce the use of knowledge graphs to measure the distance between various medical concepts and propose a novel relevance measure for the evaluation of CBIR by defining an approximate matching-based relevance score between two sets of medical concepts which allows us to indirectly measure the similarity between medical images.
We quantitatively demonstrate the effectiveness and feasibility of our relevance measure using a public dataset. 
\end{abstract}

\section{Introduction}
Along with the digitalization of healthcare and significant advancements in radiology, various multimedia data like videos, images and texts stored in the Picture Archiving and Communication Systems (PACS) of hospitals are increasing faster than ever. When facing difficult cases in clinical routines, radiologists tend to look at previous cases to determine a diagnosis. However, most access to such data are based on patient identification or keyword-based queries (such as modality, organ, symptom, etc) \cite{Henning2004benefits}which limit the potential usage of these data. Sometimes it can be hard to summarize certain cases simply using few keywords. To reduce the inaccuracy of text-based search and to fully exploit the intrinsic value of the accumulated data, there is a growing need to build efficient content-based image retrieval (CBIR) systems which accept an image as a query and returns a set of similar cases as references for doctors.


Content-Based Image Retrieval (CBIR) is a technique for retrieving images from large databases based on the visual content rather than metadata like keywords or descriptions. 
Performance evaluation plays a crucial role for CBIR systems \cite{Henning2001Performance} since it enables the comparison of different systems and allows for analysis of how these systems perform under various application scenarios. 
However, there is still a lack of universally accepted benchmarks for CBIR, making it difficult to compare different systems objectively. 
Most of the existing evaluation methods so far rely on synthetically produced ground truth or manual annotations. 
Generally the goal is to identify whether the retrieved images are relevant or not by comparing if the retrieved images share the same label as the query image. 
In fact, such a binary defined relevance remains sub-optimal since images are very complex information carriers that convey much more information compared with single labels. 
Therefore, some researchers have proposed to evaluate the performance of CBIR systems using local semantic concepts contained in images 
(e.g., ``tumors'', ``bones'', ``vessels''), since the semantic content of concepts is richer than that of single labels \cite{VOGEL2006Performance,0Integrating,2025Semantic,serieysTextguidedVisualRepresentation2022,wei2025relaxing,WEI2025104403}. 
However, only a few of them \cite{happier,Camille2014HSBD} take the subtle relationships between semantic concepts into consideration when measuring similarity between images. 
In this case, all the concepts are equally treated as independent. 
Sometimes semantically close concepts such as ``cat'' and ``feline'' are regarded as totally isolated concepts. 

In this work, we aim to design a semantically-aware relevance measure for images that can be integrated into the evaluation of CBIR tasks.
We anchor this work here in the thematic field of medical imaging but our study remains generic and generalizable to other fields.
To this end, we use external Knowledge Graphs (KGs) to model the complex relationships between semantic concepts (e.g., ``artery'', ``vessel'') which can be extracted from the descriptive text of medical images (e.g., radiological
case reports or descriptive captions in the literature). 
We use the shortest path between each concept to calculate the distance between various concepts and define a non-binary relevance measure between images based on approximate matching. This measure is finally combined with the Normalized Discounted Cumulative Gain (NDCG) method which is widely accepted in the broader field of information retrieval. 


This article is organized as it follows.
Sec.~\ref{SEC:Related works} introduces existing methods for CBIR evaluation.
In Sec.~\ref{SEC:Proposed method}, we present our contribution aiming at evaluating CBIR performance using semantically-aware relevance for medical images with the integration of external knowledge.  
In Sec.~\ref{SEC:Experiments}, an experimental evaluation is proposed both for the proposed relevance measure (named nn-IoU) and its computational cost in the context of CBIR. 
Finally, in Sec.~\ref{SEC:Conclusion}, some discussions and research perspectives are provided.

\section{Related works}
\label{SEC:Related works}

Evaluating content-based image retrieval (CBIR) systems remains a significant challenge.
As a user-oriented tool, the ultimate performance measure should always be user satisfaction. But apparently such subjective measure is not only hard to define but also vary greatly between different users and applications. Some researchers aimed to design interactive methods to account for user feedback but such works are inherently limited by perception subjectivity \cite{2003Relevance,2018Overview}. 
Therefore, objective and standardized evaluation metrics have been widely regarded as the most reliable and fair means of performance measurement. In the following of this section, we discuss some label-free(unsupervised) metrics and some other metrics supervised by either single label or multi-labels in the literature.

Some unsupervised metrics tend to generate synthetic queries by applying data augmentation techniques (e.g., rotation, noise) to original images and evaluate whether the system could retrieve the original (untransformed) image or similar images \cite{grigorova2007content}. 
But such synthetic queries may not reflect real user intents and the lack of standardization makes it hard to compare results across studies.
Other label-free metrics often aim to quantify the retrieval performances by measuring the distance between query images and retrieved images based on low-level features like colors and pixels intensity \cite{jeyakumar2015performance}. 
However, such unsupervised metrics focus on low-level visual similarity and fail to capture semantic relevance. 
As a result, supervised metrics are generally preferred for evaluating CBIR systems by leveraging labeled data, especially in domains like medical imaging where semantic understanding 
are critical.

Since retrieval and classification tasks share many similarities in common, many classical metrics for image classification have been adapted to CBIR tasks like precision and recall \cite{Henning2001Performance}. 
However, both measures are imperfect since images contain a vast amount of diverse visual information including color, texture and object composition. Precision and recall may not fully capture the nuances of these features using only a single label. 
Although measures like F1-score and mean average precision have also been proposed to balance precision and recall and to account for ranking order, such measures commonly used in classification tasks remains sub-optimal because classification tasks always deal with a limited number of categories. In fact, retrieval tasks do not deal with fixed classes of items and the real intents of the users can be diverse and complex. For instance, if a user searches for "a person riding a bicycle in a park on a rainy night", measures like precision and recall are unlikely to fulfill the user's intent because such a label rarely exists in any dataset.
Furthermore, classification metrics often presume that all retrieved images can be binarily regarded as either relevant or irrelevant simply by comparing if retrieved images share the same label as the query image.
In fact, since an image contains much more information than a single label, relevance between images should take into account non-binary similarity if possible (e.g., highly relevant, somewhat relevant, irrelevant).

Instead of using a single label, some researchers have discussed how to measure the relevance of images based on multiple semantic concepts contained in images \cite{VOGEL2006Performance,serieysTextguidedVisualRepresentation2022}. 
Vogel et al. \cite{VOGEL2006Performance} proposed to divide one image into 10$\times{}$10 patches and to detect local objects/concepts contained in each image patch separately. 
In this case, an image can be represented, for example, as 40\% sky + 30\% grass + 30\% buildings. 
Thus, the relevance between images can be determined by a combination of semantic concepts instead of single labels. 
Serieys et al. \cite{serieysTextguidedVisualRepresentation2022} proposed to represent a medical image using semantic concepts extracted from the descriptive text of that image, and they used the Intersection over Union (IoU) between two sets of concepts to measure the non-binary relevance between medical images. 
However, such a measure remains imperfect because it neglects the subtle relationships between semantic concepts (e.g., ``vessel'' and ``artery'' which represent very close concepts, can be treated as totally independent terms). 
Meanwhile, such common issues can be solved by approximate matching.

Approximate matching \cite{breitinger2014approximate} refers to a data processing technique often used to find strings that approximately match a given pattern, rather than exactly match it. 
This is particularly useful when dealing with data that may contain typos, variations in spelling with the integration of Levenshtein Distance which measures the number of single character edits (insertions, deletions, or substitutions) required to change one string into the other. 
Similarly, such a approximate matching strategy could also be used to distinguish close but unique medical concepts like ``vein'' and ``vessel'' using specific similarity measures.
In this case, Knowledge Graphs (KG) are crucial for measuring concept similarity because they provide a structured and semantic representation of relationships with rich context. They enable accurate, interpretable, and domain-specific similarity measures. Various methods have been discussed to measure node similarity using ontologies \cite{zheng2019taxonomy}, e.g., path-based methods tend to measure the similarity of two nodes based on the shortest path between them or their closest common ancestor \cite{happier}. Graph embedding-based methods tend to directly calculate the cosine similarity of two nodes after representing the nodes as vectors using graph embedding techniques (e.g., Node2Vec \cite{node2vec}, TransE \cite{bordesTranslatingEmbeddingsModeling}).
In this work, we propose a semantically-aware measure for CBIR evaluation by introducing external KG to better quantify the relevance between images using semantic concepts extracted from the descriptive text of corresponding images.


\section{Proposed measure: nn-CUI@K}
\label{SEC:Proposed method}

In this section, we will present the underlying methodological concepts of the proposed measure (named nn-CUI@K) and details of its implementation. 
Given a dataset of medical image-caption pairs, we first extract medical concepts, labeled as Concept Unique Identifiers (CUIs), from the descriptive text of the corresponding image. 
From these CUI-labeled images, we then introduce an approach relying on external KG to measure the distance between each CUI and further compute the relevance between images based on approximate matching for two sets of CUIs. 

\subsection{CUI (concept) extraction}
The CUI is the basic component of the Unified Medical Language System (UMLS) \cite{bodenreiderUnifiedMedicalLanguage2004} which can be regarded as a huge vocabulary encompassing a wide range of medical concepts.
In the UMLS, each medical concept is assigned a CUI, consisting of the letter "C" followed by seven digits (e.g., the CUI C0042449 refers to the concept \textit{Veins}).

For each image in a medical image-caption pair dataset, we first extract CUIs from the corresponding descriptive text using Named Entity Recognition tools like MedCAT \cite{Kraljevic2021MedCat}. 
Such a prerequisite has already been done for some datasets (e.g., ROCO \cite{pelkaRadiologyObjectsCOntext2018}). 

\begin{figure}[!t]
\centerline{\includegraphics[width=\linewidth]{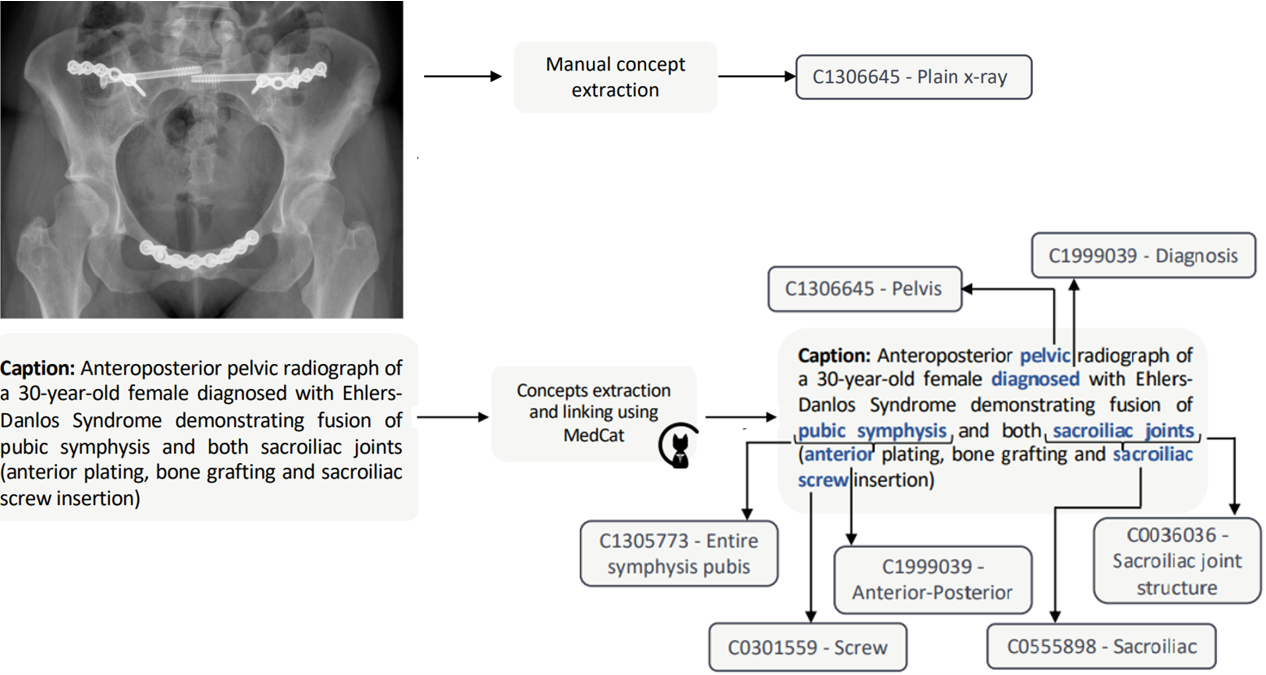}}
\caption{CUI extraction in the ROCOv2 dataset \cite{Johannes0ROCOv2}.}
\label{fig1}
\end{figure}

\subsection{Distance measurement between CUIs}
Previous measures proposed in the literature often neglect the relatedness between CUIs.
However, some semantically close CUIs may be regarded as totally independent or even isolated CUIs (e.g., ``C0042449:veins'' and ``C0005847:blood vessels''). 
To avoid such misleading judgments when computing relevance between medical images using CUIs, we choose the KG stored in the UMLS to model the complex relations between CUIs and to measure the distance between CUIs in the next steps. 
Since the KG contains comprehensive semantic networks of medical domain, we choose to measure the distance between CUIs $x$ and $y$ simply according to the length of the shortest path connecting the two CUIs:
\begin{equation}
    dist\left ( x,y \right )=len\left (shortest\_path(x,y) \right )
\end{equation}
e.g., for directly connected CUIs in the KG like \textit{C0042449: veins} and \textit{C0005847:blood vessels}, the distance $dist\left ( C0042449,C0005847 \right ) =1$. 
For not directly connected CUIs like \textit{C0006121:brain stem} and \textit{C0018670:head}, $dist\left ( C0006121,C0018670\right )=3$ since the shortest path between them is ``\textit{C0006121}\textbf{-}\textit{C0006104}\textbf{-}\textit{C0926510}\textbf{-}\textit{C0018670}''.

It needs to be noticed that we chose a directed acyclic graph to compute the distances and other measures can also be considered as mentioned in Sec.~\ref{SEC:Related works}.


\subsection{Semantic-aware relevance measure: nn-IoU}
\label{sec:nn-iou}


Some measures have been used to evaluate the similarity between two sets of medical concepts such as IoU proposed by \cite{serieysTextguidedVisualRepresentation2022} and Bipartite Matching (BM) proposed in \cite{zheng2019taxonomy}. 
However, IoU only considers the intersection of two sets of concepts as relevant and overlooks the cases where different concepts can also be extremely similar (e.g., ``C0042449:veins'' and ``C0005847:blood vessels''). 
On the other hand, while BM takes into account the relatedness between each CUI, it fails to notice the varying degree of relevance especially when CUI set $A$ is a subset of CUI set $B$. 
Considering the above limitations, we propose a novel measure called nearest neighbor-based Intersection over Union (nn-IoU):
\begin{equation}
\text{nn-IoU}\left ( A,B \right )=\frac{\left|A\cap B\right|+\lambda *\left|rel(A,B)\right|}{\left|A\cup B\right|}   \label{eq:nn-iou}
\end{equation}
where $A\cup B$ and $A\cap B$ refer to the total CUIs and the overlapped CUIs for CUI sets $A$ and $B$. 
Besides, $rel(A,B)$ denotes the (directly) connected CUIs (also called neighbours) between CUI sets $A$ and $B$, which can be computed as shown in the algorithm~\ref{alg:nn-IoU}. 
$\lambda$ is a coefficient between zero and one to balance the importance for identical overlapping CUIs and the directly related CUIs. 


\begin{algorithm}
  \caption{Identifying related concepts}
  \label{alg:nn-IoU}
  \KwIn{CUI sets A $\&$ B\;  \,\,\,\,\,\,\,\,\,\,\,\,\,\,\,\,
  distance threshold $n$}
  \KwOut{$rel(A,B)$}
  \textcolor{blue}{$\#$ Initialize $rel(A,B)$ as an empty set\;}
  $rel(A,B)=\emptyset$\;
  \For{$a \in A$}
  {
    \For{$b \in B$}
    {
      \textcolor{blue}{$\#$ Select nearest neighbours\;}
      \If{$dist(a,b) \le$ threshold $n$} 
      {
      \textcolor{blue}{$\#$ No CUI appear twice in the numerator\;}
      \If{$(a \notin A\cap B) \wedge (a \notin rel(A,B))$}
      {
      add $a$ to $rel(A,B)$
      }
      \If{$(b \notin A\cap B) \wedge (b \notin rel(A,B))$}
      {
      add $b$ to $rel(A,B)$
      }
      }
    }
  }
  return $rel(A,B)$\;
\end{algorithm}


\subsection{Evaluation based on ranking order}
Based on the relevance defined in Sec.~\ref{sec:nn-iou}, we finally compute our proposed semantically-aware measure nn-CUI@K by considering the following steps:
\begin{enumerate}
    \item For each image in the test set, we extract CUIs from its descriptive caption;
\item With each image as a query, we first compute the relevance score using nn-IoU as defined in  Sec.\ref{sec:nn-iou} between the set of CUIs paired with that image and the set of
CUIs paired with all candidate images. We then rank all candidate
images in descending order based on nn-IoU scores and treat
this ranking as the ground truth for the image retrieval task; 
\item We first compute the sum of the discounted relevance scores for the top $K$ retrieved images, called the Discounted Cumulative Gain (DCG). 
We then compute the Ideal DCG (IDCG) score using the ground truth ranking;
\item The final score is the average of the Normalized Discounted Cumulative Gain (NDCG) values over all query images.
\end{enumerate}

Specifically, given a test set of $N$ medical image-text pairs and a CBIR system $M$, the nn-CUI@K score over this dataset can be computed as detailed in Algorithm~\ref{alg:nn-CUI@K}.

\begin{algorithm}[t]
\caption{Implementation of nn-CUI@K}
\label{alg:nn-CUI@K}
\KwIn{medical image-text pair dataset $\left \{ V_i,T_i \right \}_{i=1}^N$ \;\,\,\,\,\,\,\,\,\,\,\,\,\,\,\,\,\,\,CBIR system M}
\KwOut{nn-CUI@K}

Extract CUIs from $\{ V_i,T_i \}_{i=1}^N$\; 
$NDCG \leftarrow 0$\;

\For{$i \leftarrow 1$ \KwTo $N$}{
    $DCG \leftarrow 0$, $IDCG \leftarrow 0$\;
    
    Use $V_i$ as query image\;
    
    Get top-$K$ similar images $\{ r_j \}_{j=1}^K$ using system $M$\;
    
    \For{$j \leftarrow 1$ \KwTo $K$}{
        $relevance \leftarrow \text{nn-IoU}(CUI(r_j), CUI(V_i))$\;
        
        $penalty \leftarrow \log_2(j+1)$\;
        
        $DCG \leftarrow DCG + \frac{relevance}{penalty}$\;
    }
    
    Get top-$K$ similar images $\{ R_j \}_{j=1}^K$ using nn-IoU\;
    
    \For{$j \leftarrow 1$ \KwTo $K$}{
        $relevance \leftarrow \text{nn-IoU}(CUI(R_j), CUI(V_i))$\;
        
        $penalty \leftarrow \log_2(j+1)$\;
        
        $IDCG \leftarrow IDCG + \frac{relevance}{penalty}$\;
    }
    
    $NDCG \leftarrow NDCG + \frac{DCG}{IDCG}$\;
}
\Return{$\frac{NDCG}{N}$}
\end{algorithm}



    





\section{Experimental study}
\label{SEC:Experiments}
The only difference between the proposed CBIR evaluation metric nn-CUI@K and the CUI@K proposed in \cite{serieysTextguidedVisualRepresentation2022} is that we compute the relevance of medical images using nn-IoU instead of IoU. 
To showcase the interest of nn-IoU, we evaluate its performance in the context of CUI-based image retrieval in the following experiments.

\subsection{Datasets}
ROCOv2 is a multimodal dataset \cite{Johannes0ROCOv2} 
 consisting of 60,163/9948/9928 images in the training, validation and test splits. All the images are extracted from PubMed \cite{canese2013pubmed} articles with corresponding captions covering a wide range of modalities and organs. 
Besides, each image is paired with a set of CUIs extracted from its caption. 


\subsection{Implementation details}

Only hierarchical relations (i.e., is\_a relations) for the KG from the UMLS metathesaurus are used to measure the distance between each CUI. 
The length of the shortest paths between all involved CUIs are precomputed offline using NetworkX \cite{hagberg2008exploring}.
The distance threshold $n$ is set to 1. 
The $\lambda$ value for nn-IoU score is set to 0.5.
We justify the hyperparameter settings in the following ablation study.

\subsection{Evaluation for relevance measure nn-IoU}

To quantitatively evaluate the effectiveness of the proposed relevance measure nn-IoU vs. IoU, we perform image retrieval on ROCOv2, since a well-designed relevance measure should help to identify the most relevant images for a given query. 

Specifically, we first perform image retrieval based only on the corresponding CUIs using nn-IoU $\&$ IoU as the relevance measure. 
Additionally we also perform image retrieval based solely on textual and visual content instead of discrete CUIs. 
To do so, we encode each image-text pair using the BioMedCLIP visual and textual encoders \cite{BiomedCLIP} respectively, and then retrieve the most similar images based on the cosine similarity of the encoded visual and textual embeddings.

We then use Precision@K to evaluate the retrieval performance for each relevance measure. 
Categorical modality/organ labels are obtained by mapping each class to corresponding CUIs. 
The experimental scores are presented in Table~\ref{tab:score}.
We can observe that nn-IoU outperforms IoU in all the tasks and the relatively small performance gap may stem from the nature of ROCO dataset (each image is paired with a short caption containing only 3.4 CUIs in average such that the potential relatedness may not be ubiquitous).

We also observe that nn-IoU leads to an even better retrieval result compared to the retrieval performance of both textual and visual encoders of BioMedCLIP (except Precision@5 score for modality\&organ) which confirms its superiority in measuring the relevance of medical images using CUIs.

\begin{table}[]
\caption{Evaluation of image retrieval using various relevance measures on the ROCOv2 test set. 
BioMedCLIP-v/t refer to visual and textual embeddings encoded using BioMedCLIP.}
\centering
\begin{tabular}{|l|lllllllll|}
\hline
\multirow{3}{*}{measure}     & \multicolumn{9}{c|}{Precision@K}                                                                                               \\
                            & @5    & @10   & @30                        & @5    & @10   & @30                        & @5         & @10        & @30        \\
                            & \multicolumn{3}{c}{Modality}               & \multicolumn{3}{c}{Organ}                  & \multicolumn{3}{c|}{Modality\&Organ} \\ \hline
IoU                         & 99.28 & 99.17 & \multicolumn{1}{l|}{98.99} & 94.27 & 93.46 & \multicolumn{1}{l|}{91.15} & 90.03      & 88.87      & 85.72      \\ \cline{1-1}\hline
\textbf{nn-IoU}                      & \textbf{99.34} & \textbf{99.46} & \multicolumn{1}{l|}{\textbf{99.42}} & \textbf{94.59} & \textbf{93.78} & \multicolumn{1}{l|}{\textbf{91.50}} & 90.65      & \textbf{89.53}      & \textbf{86.85}      \\ \cline{1-1}\hline
BioMedCLIP-t                & 74.61 & 73.25 & \multicolumn{1}{l|}{69.83} & 60.52 & 56.91 & \multicolumn{1}{l|}{51.17} & 37.62      & 33.53      & 27.44      \\ \cline{1-1}\hline
BioMedCLIP-v                & 95.85 & 93.09 & \multicolumn{1}{l|}{89.44} & 87.12 & 84.32 & \multicolumn{1}{l|}{78.46} & \textbf{91.33}      & 89.48      & 84.08      \\ \cline{1-1}\hline
\end{tabular}
\label{tab:score}
\vspace{-3mm}
\end{table}

\subsection{Ablation study}

\begin{figure}[!t]
\centerline{\includegraphics[width=0.5\linewidth]{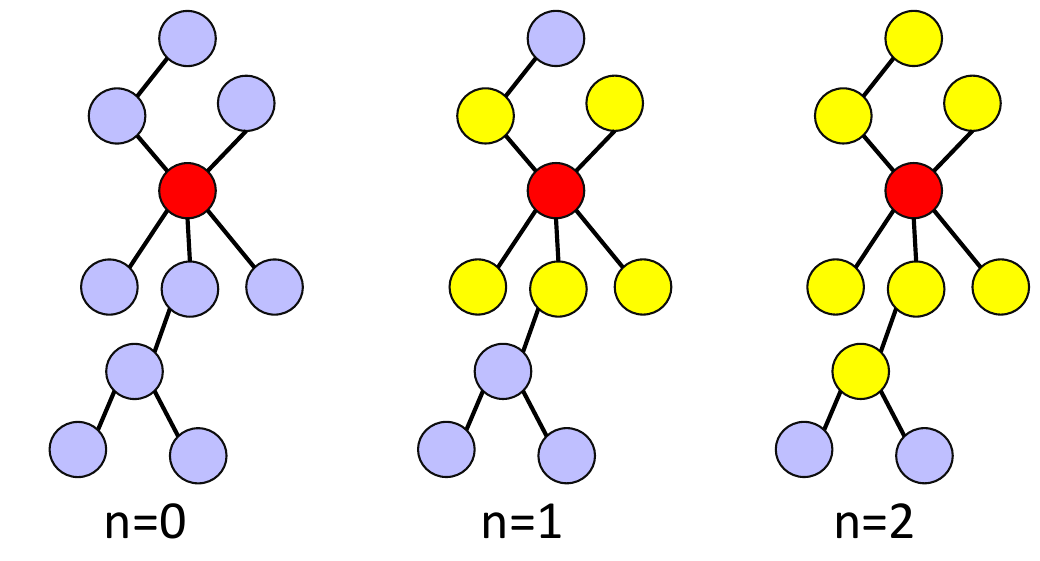}}
\caption{Ablation study of including more neighbours of CUIs for approximate matching by increasing $n$ from 0 to 2 where $n$ denotes the CUI distance threshold. Yellow nodes  correspond to the nearest neighbours of the red node.}
\label{fign}
\vspace{-4mm}
\end{figure}

\begin{figure}[!t]
\centerline{\includegraphics[width=0.73\linewidth]{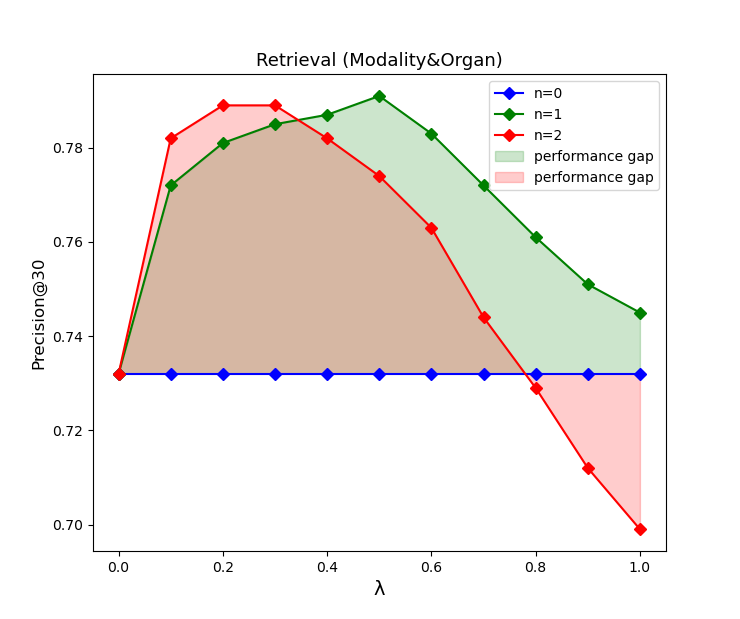}}
\vspace{-3mm}
\caption{P@30 scores in the ablation study. 
The retrieval precision will keep increasing until $\lambda=0.5$ and $\lambda=0.2$ when increasing $\lambda$ from 0 to 1 for $n=1$ and $n=2$ respectively.}
\label{n012}
\vspace{-4mm}
\end{figure}

As shown in the Equation \ref{eq:nn-iou}, the major difference between IoU and nn-IoU is the newly added term $rel(A,B)$ in the numerator which accounts for the approximate matching by matching each CUI with not only itself but also with its nearest neighbours in the KG.
To identify if it is beneficial to add such a term, we aim to quantify the impact of $rel(A,B)$ which enables approximate matching. Therefore, in the following ablation study we freeze all the parameters for nn-IoU except the distance threshold $n$ which controls the number of nearest neighbours and the coefficient $\lambda$ which balances the weight of $rel(A,B)$.
To highlight the impact of the approximate match, we perform image retrieval tasks on an isolated subset of ROCOv2 where IoU and nn-IoU exhibit inconsistent performance as shown in the Fig.~\ref{fign}.
The P@30 experimental scores are shown in Fig.~\ref{n012}. 
As it can be observed, when $n=0$, no nearest neighbour will be counted and in this case nn-IoU is equivalent to IoU, which explains why there is no variation by increasing $\lambda$. And the areas under curves indicate the performance gap between Iou (n=0) and nn-IoU (n=1 \& n=2).
We can also observe that the green and red curves are always above the blue one when $\lambda \in  \left[0,0.7\right]$, which highlights that it is beneficial to add $rel(A,B)$ paired with an appropriate weight. 
We also find that it could be detrimental to assign a higher value for $\lambda$ especially when $n=2$ as some noise might also be included with a relatively high weight, leading to inaccurate relevance. 

\subsection{Time cost of evaluation}
\begin{figure}[!t]
\centerline{\includegraphics[width=0.75\linewidth]{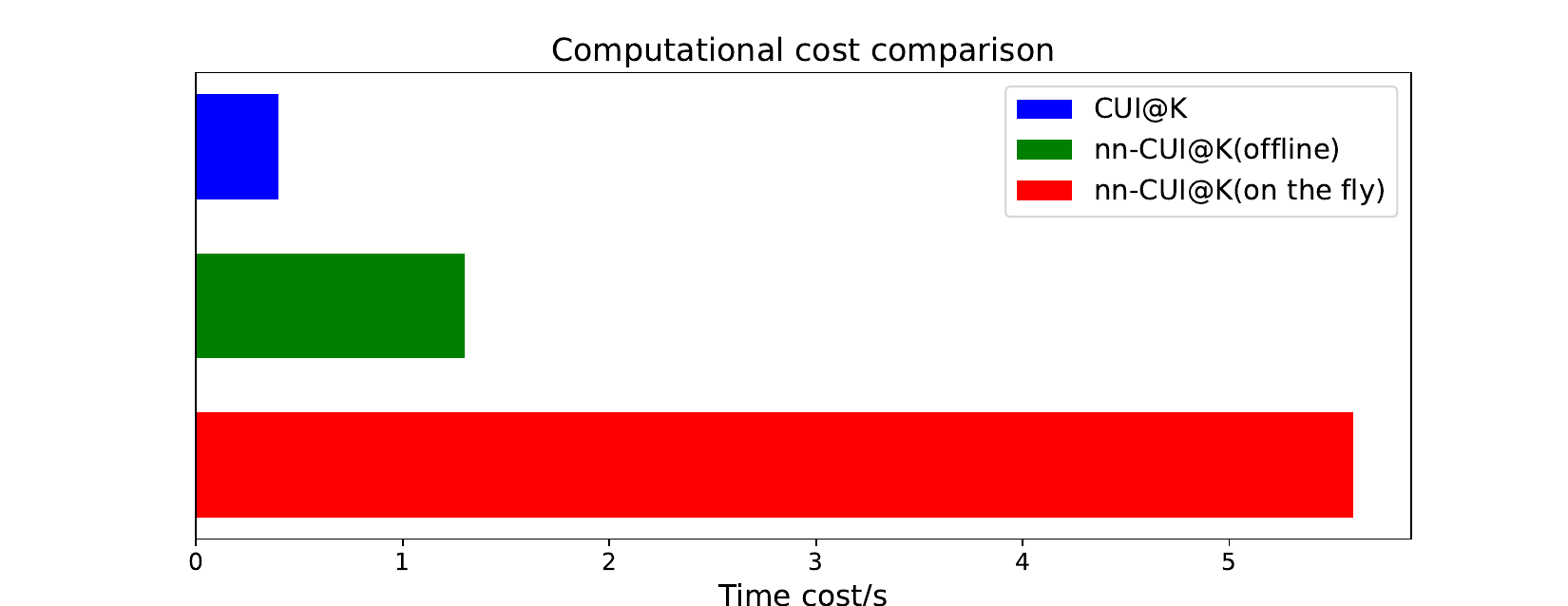}}
\caption{Average computational cost per image on ROCOv2.}
\label{time}
\vspace{-3mm}
\end{figure}

Compared to IoU, our relevance measure nn-IoU inevitably comes with a higher computational cost by introducing an additional approximate matching for CUIs expressed as term $rel(A,B)$. 
Since UMLS models the relations between all CUIs using a huge KG, it can be computationally expensive to find the potential nearest neighbours for CUIs by searching for the shortest path between two CUIs. 
Instead of computing the distances between the individual CUIs on the fly, we propose to compute the necessary distances between the CUIs off-line and to store the nearest neighbours of each CUI in a dictionary. 
In this case we can avoid the replication computation and $rel(A,B)$ can be computed even without KG. 
We provide the evaluation time cost on ROCOv2 in Fig.~\ref{time}.


\section{Conclusion}
\label{SEC:Conclusion}
In this work, we propose a KG-augmented evaluation measure nn-CUI@K for CBIR by replacing the IoU which is used in CUI@K \cite{serieysTextguidedVisualRepresentation2022} with a novel semantically-aware relevance measure called $nn-IoU$, where we add a new term to account for potential relatedness between CUIs. Based on experiments on ROCOv2, we find that, as a relevance measure, nn-IoU leads to  more accurate retrieval results compared to IoU highlighting the effectiveness of approximate matching based on nearest neighbours. 
On the other hand, we believe that our measure could also be extended to other applicative domains using professional KGs like WordNet \cite{miller1995wordnet} and ENVO \cite{buttigieg2013environment}. However, the proposed measure still has some limitations offering extensional research directions, e.g., only the lengths of shortest paths between CUIs are taken into consideration to detect similar concepts in nn-IoU while all other CUIs are treated as irrelevant. 
Besides, other information like the depth of nodes in the taxonomy can also help to measure the similarity between CUIs. Additionally, our method also comes with higher computational cost, though the shortest path computation can be pre-computed off-line to speed up the evaluation procedure, scalability remains another concern when evaluating on larger datasets using larger KGs.


%
%
\bibliographystyle{splncs04}
\bibliography{main}

\begin{thebibliography}{10}
\providecommand{\url}[1]{\texttt{#1}}
\providecommand{\urlprefix}{URL }
\providecommand{\doi}[1]{https://doi.org/#1}

\bibitem{bodenreiderUnifiedMedicalLanguage2004}
Bodenreider, O.: The {{Unified Medical Language System}} ({{UMLS}}): Integrating biomedical terminology. Nucleic Acids Research pp. 267D--270 (2004)

\bibitem{bordesTranslatingEmbeddingsModeling}
Bordes, A., Usunier, N., {Garcia-Duran}, A., Weston, J., Yakhnenko, O.: Translating {{Embeddings}} for {{Modeling Multi-relational Data}}. In: NIPS, Procs. pp. 2787--2795 (2013)

\bibitem{breitinger2014approximate}
Breitinger, F., Breitinger, F., White, D., Guttman, B., McCarrin, M., Roussev, V.: Approximate matching: definition and terminology. US Department of Commerce, National Institute of Standards and Technology (2014)

\bibitem{buttigieg2013environment}
Buttigieg, P.L., Morrison, N., Smith, B., Mungall, C.J., Lewis, S.E., Consortium, E.: The environment ontology: contextualising biological and biomedical entities. Journal of biomedical semantics  \textbf{4}, ~1--9 (2013)

\bibitem{canese2013pubmed}
Canese, K., Weis, S.: Pubmed: the bibliographic database. The NCBI handbook  \textbf{2}(1) (2013)

\bibitem{grigorova2007content}
Grigorova, A., De~Natale, F.G., Dagli, C., Huang, T.S.: Content-based image retrieval by feature adaptation and relevance feedback. IEEE transactions on multimedia  \textbf{9}(6),  1183--1192 (2007)

\bibitem{node2vec}
Grover, A., Leskovec, J.: Node2vec: Scalable feature learning for networks. In: SIGKDD, Procs. pp. 855--864 (2016)

\bibitem{hagberg2008exploring}
Hagberg, A., Swart, P.J., Schult, D.A.: Exploring network structure, dynamics, and function using networkx. Tech. rep., Los Alamos National Laboratory (LANL), Los Alamos, NM (United States) (2008)

\bibitem{jeyakumar2015performance}
Jeyakumar, V., Kanagaraj, B.R.: Performance evaluation of image retrieval system based on error metrics. Indian journal of science and technology  \textbf{8}, ~117 (2015)

\bibitem{zheng2019taxonomy}
Jia, Z., Lu, X., Duan, H., Li, H.: Using the distance between sets of hierarchical taxonomic clinical concepts to measure patient similarity. {BMC} Medical Informatics Decis. Mak.  \textbf{19}(1),  91:1--91:11 (2019)

\bibitem{Kraljevic2021MedCat}
Kraljevic, Z., Searle, T., Shek, A., Roguski, L., Noor, K., Bean, D., Mascio, A., Zhu, L., Folarin, A.A., Roberts, A., Bendayan, R., Richardson, M.P., Stewart, R., Shah, A.D., Wong, W.K., Ibrahim, Z.M., Teo, J.T., Dobson, R.J.B.: Multi-domain clinical natural language processing with medcat: The medical concept annotation toolkit. Artif. Intell. Medicine  \textbf{117},  102083 (2021)

\bibitem{Camille2014HSBD}
Kurtz, C., Beaulieu, C.F., Napel, S., Rubin, D.L.: A hierarchical knowledge-based approach for retrieving similar medical images described with semantic annotations. J. Biomed. Informatics  \textbf{49},  227--244 (2014)

\bibitem{miller1995wordnet}
Miller, G.A.: Wordnet: a lexical database for english. Communications of the ACM  \textbf{38}(11),  39--41 (1995)

\bibitem{Henning2004benefits}
M{\"{u}}ller, H., Michoux, N., Bandon, D., Geissb{\"{u}}hler, A.: A review of content-based image retrieval systems in medical applications - clinical benefits and future directions. Int. J. Medical Informatics  \textbf{73}(1),  1--23 (2004)

\bibitem{2018Overview}
Muramatsu, C.: Overview on subjective similarity of images for content-based medical image retrieval. Radiological Physics and Technology  \textbf{11}(1) (2018)

\bibitem{Henning2001Performance}
Müller, H., Müller, W., Squire, D.M., Marchand-Maillet, S., Pun, T.: Performance evaluation in content-based image retrieval: overview and proposals. Pattern Recognition Letters (5),  593--601 (2001)

\bibitem{pelkaRadiologyObjectsCOntext2018}
Pelka, O., Koitka, S., R{\"u}ckert, J., Nensa, F., Friedrich, C.M.: Radiology {{Objects}} in {{COntext}} ({{ROCO}}): {{A Multimodal Image Dataset}}. In: Intravascular {{Imaging}} and {{Computer Assisted Stenting}} and {{Large-Scale Annotation}} of {{Biomedical Data}} and {{Expert Label Synthesis}}, vol. 11043, pp. 180--189. {Springer International Publishing}, {Cham} (2018). \doi{10.1007/978-3-030-01364-6_20}

\bibitem{happier}
Ramzi, E., Audebert, N., Thome, N., Rambou, C., Bitot, X.: Hierarchical average precision training for pertinent image retrieval. In: ECCV, Procs. (2022)

\bibitem{Johannes0ROCOv2}
Rückert, J., Bloch, L., Brüngel, R., Idrissi-Yaghir, A., Schfer, H., Schmidt, C.S., Koitka, S., Pelka, O., Abacha, A.B., Herrera, A.G.S.D.: Rocov2: Radiology objects in context version 2, an updated multimodal image dataset. Scientific Data

\bibitem{serieysTextguidedVisualRepresentation2022}
Serieys, G., Kurtz, C., Fournier, L., Cloppet, F.: Text-guided visual representation learning for medical image retrieval systems. In: ICPR, Procs. pp. 593--598 (2022)

\bibitem{2025Semantic}
Vagena, Z., Wei, X., Kurtz, C., Cloppet, F.: Semantic aware representation learning for optimizing image retrieval systems in radiology. Pattern Recognition  \textbf{158} (2025)

\bibitem{VOGEL2006Performance}
Vogel, J., Schiele, B.: Performance evaluation and optimization for content-based image retrieval. Pattern Recognit.  \textbf{39}(5),  897--909 (2006)

\bibitem{WEI2025104403}
Wei, X., Kurtz, C., Cloppet, F.: Enhancing vision-language contrastive representation learning using domain knowledge. Computer Vision and Image Understanding p. 104403 (2025). \doi{https://doi.org/10.1016/j.cviu.2025.104403}, \url{https://www.sciencedirect.com/science/article/pii/S1077314225001262}

\bibitem{wei2025relaxing}
Wei, X., Kurtz, C., Cloppet, F.: Relaxing binary constraints in contrastive vision-language medical representation learning. In: 2025 IEEE/CVF Winter Conference on Applications of Computer Vision (WACV). pp. 4462--4471. IEEE (2025)

\bibitem{0Integrating}
Wei, X., Vagena, Z., Kurtz, C., Cloppet, F.: Integrating expert knowledge with vision-language model for medical image retrieval. In: ISBI, Procs. pp. 897--909 (2024)

\bibitem{BiomedCLIP}
Zhang, S., Xu, Y., Usuyama, N., Bagga, J., Tinn, R., Preston, S., Rao, R., Wei, M., Valluri, N., Wong, C., Lungren, M.P., Naumann, T., Poon, H.: Large-scale domain-specific pretraining for biomedical vision-language processing. CoRR  \textbf{abs/2303.00915} (2023)

\bibitem{2003Relevance}
Zhou, X.S., Huang, T.S.: Relevance feedback in image retrieval: A comprehensive review. Multimedia Systems  \textbf{8}(6),  536--544 (2003)

\end{thebibliography}
\end{document}